\documentclass{article}

\PassOptionsToPackage{numbers}{natbib}

\usepackage[preprint]{neurips_2019}

\usepackage[utf8]{inputenc} 
\usepackage[T1]{fontenc}    
\usepackage{hyperref}       
\usepackage{url}            
\usepackage{booktabs}       
\usepackage{amsfonts}       
\usepackage{nicefrac}       
\usepackage{microtype}      

\usepackage{amsmath}
\usepackage{amssymb}
\usepackage[dvipsnames]{xcolor}
\usepackage{kvoptions}
\usepackage{xcolor-material}
\usepackage{algorithm}
\usepackage{algorithmicx}
\usepackage{algpseudocode}
\usepackage{mathabx}
\usepackage{dsfont}
\usepackage{mathtools}

\usepackage{tikz}
\usepackage{pgfplots}

\newtheorem{definition}{Definition}
\newtheorem{example}{Example}

\newcommand{\pl}{\mathcal{P}}

\newcommand{\expe}{\mathbb{E}}
\newcommand{\hi}{H}
\newcommand{\transfunc}{\mathcal{T}}
\newcommand{\obsfunc}{\mathcal{O}}
\newcommand{\scorefunc}{J}
\newcommand{\defeq}{\vcentcolon=}

\newcommand{\eg}{e.g.}
\newcommand{\ie}{i.e.}

\newcommand{\E}[2]{\operatorname{\mathbb{E}}_{#1}\left[#2\right]}

\DeclareMathOperator*{\argmax}{arg\,max}

\newcommand{\ent}{\mathcal{H}}

\newcommand{\voidarg}{{\,\cdot\,}}

\newcommand{\sspace}{\mathcal{S}}
\newcommand{\aspace}{\mathcal{A}}

\newcommand{\state}{s}

\newcommand{\st}{{\state_t}}

\newcommand{\stp}{{\state_{t+1}}}

\newcommand{\action}{a}

\newcommand{\at}{{\action_t}}

\newcommand{\reward}{r}

\newcommand{\rt}{\reward_t}

\newcommand{\Q}{Q}

\newcommand{\A}{\mathcal{A}}

\newcommand{\policy}{\pi}

\newcommand{\params}{\theta}

\newcommand{\pparams}{{\phi}}   
\newcommand{\qparams}{{\theta}}   
\newcommand{\vparams}{{\psi}}   
\newcommand{\vtargetparams}{{\bar\psi}}   

\title{Coordination in Adversarial Sequential Team Games via Multi-agent Deep Reinforcement Learning}

\author{%
  Andrea Celli\\
  Politecnico di Milano\\
  \texttt{andrea.celli@polimi.it} \\
  \And
  Marco Ciccone\\
  Politecnico di Milano\\
  \texttt{marco.ciccone@polimi.it} \\
  \And
  Raffaele Bongo\\
  Politecnico di Milano\\
  \texttt{raffaele.bongo@mail.polimi.it} \\
   \And
  Nicola Gatti\\
  Politecnico di Milano\\
  \texttt{nicola.gatti@polimi.it} \\
}

\begin{document}

\maketitle

\begin{abstract}
    Many real-world applications involve teams of agents that have to
    coordinate their actions to reach a common goal against potential
    adversaries. This paper focuses on zero-sum games where a team of players
    faces an opponent, as is the case, for example, in Bridge, collusion in
    poker, and collusion in bidding. The possibility for the team members to
    communicate before gameplay---that is, coordinate their strategies ex
    ante---makes the use of behavioral strategies unsatisfactory. We introduce
    Soft Team Actor-Critic (STAC) as a solution to the team's coordination
    problem that does not require any prior domain knowledge. STAC allows team
    members to effectively exploit ex ante communication via exogenous signals
    that are shared among the team. STAC reaches near-optimal coordinated
    strategies both in perfectly observable and partially observable games,
    where previous deep RL algorithms fail to reach optimal coordinated
    behaviors.
\end{abstract}

\section{Introduction}

In many strategic interactions agents have similar goals and have incentives to team up, and share their final reward.
In these settings, coordination between team members plays a crucial role.
The benefits from team coordination (in terms of team's expected utility) depend on the communication possibilities among team members.
This work focuses on {\em ex ante coordination}, where team members have an opportunity to discuss and agree on tactics before the game starts, but will be unable to communicate during the game, except through their publicly-observed actions.
Consider, as an illustration, a poker game where some players are colluding against some identified target players and will share the final winnings after the game.
Non-recreational applications are ubiquitous as well. This is the case, for instance, of collusion in bidding, where communication during the auction is illegal, and coordinated swindling in public.

Ex ante coordination enables the team to obtain significantly higher returns (up to a linear factor in the size of the game-tree) than the return they would obtain by abstaining from coordination~\cite{basilico2017team,celli2018computational}.

Finding an equilibrium with ex ante coordination is \textsf{NP}-hard and inapproximable~\cite{basilico2017team,celli2018computational}.
\citet{celli2018computational} introduced the first algorithm to compute optimal coordinated strategies for a team playing against an adversary.
At its core, it's a column generation algorithm exploiting a hybrid representation of the game, where team members play joint normal-form actions while the adversary employs sequence-form strategies~\cite{von2008extensive}.
More recently,~\citet{Farina&Celli2017} proposed a variation of fictitious play, named {\em Fictitious Team-Play} (FTP), to compute an approximate solution to the problem.
Both approaches require to iteratively solve {\em mixed-integer linear programs} (MILP), which significantly limits the scalability of these techniques to large problems, with the biggest instances solved via FTP being in the order of 800 infosets per player.\footnote{From a practical perspective, the hardness of the problem usually reduces to either solving an LP with an exponential number of possible actions (\ie, team's joint plans), or to solving best-response MILPs on a compact action space.}
The biggest crux of these tabular approaches is the need for an explicit representation of the sequential game, which may not be exactly known to players, or could be too big to be even stored in memory.
For this reason, extremely large games are usually abstracted by bucketing similar states together.
The problem with this approach is that abstractions procedures are domain-specific, require extensive domain knowledge, and fail to generalize to novel situations.

Popular equilibrium computation techniques such as fictious play (FP)~\cite{brown1951,robinson1951} and counterfactual regret minimization (CFR)~\cite{zinkevich2008regret} are not model-free, and therefore are unable to learn in a sample-based fashion (\ie, directly from experience).
The same holds true for the techniques recently employed to achieve expert-level poker playing, which are heavily based on poker-specific domain knowledge~\cite{moravvcik2017deepstack,brown2017safe,brown2018superhuman,brown2019superhuman}.
There exist model-free variations of FP and CFR~\cite{HeinrichS16,brown2018deep}.
However, these algorithms fail to model coordinated team behaviors even in simple toy problems.


On the other side of the spectrum with respect to tabular equilibrium computation techniques there are Multi-Agent Reinforcement Learning (MARL) algorithms~\cite{bu2008comprehensive,hernandez2019survey}.
These techniques don't require a perfect knowledge of the environment and are sample-based by nature, but applying them to imperfect-information sequential games presents a number of difficulties: i) hidden information and arbitrary action spaces make players non homogeneous; ii) player's policies may be conditioned only on local information.
The second point is of crucial importance since we will show that, even in simple settings, it is impossible to reach an optimal coordinated strategy without conditioning policies on exogenous signals.
Conditioning policies on the observed histories allows for some sort of implicit coordination in perfectly observable games (\ie, games where each player observes and remembers other player's actions), but this doesn't hold in settings with partial observability, which are ubiquitous in practice.

\paragraph{Our contributions}
In this paper, we propose Soft Team Actor-Critic (STAC), which is an end-to-end framework to learn coordinated strategies directly from experience, without requiring the perfect knowledge of the game-tree.
We design an ex ante signaling framework which is principled from a game-theoretic perspective, and show that team members are able to associate shared meanings to signals that are initially uninformative.
We show that our algorithm achieves strong performances on standard imperfect-information benchmarks, and on coordination games where known deep multi-agent RL algorithms fail to reach optimal solutions.

\section{Preliminaries}

In this section, we provide a brief overview of reinforcement learning and extensive-form games. For details, see~\citet{sutton2018reinforcement,shoham2008multiagent}.

\subsection{Reinforcement Learning}
At each state $s\in \sspace$, an agent selects an action $a\in\aspace$ according to a policy $\pi: \sspace\to \Delta(\aspace)$, which maps states into probability distributions over the set of available actions.
This causes a transition on the environment according to the state transition function $\transfunc:\sspace\times \aspace\to\Delta(\sspace)$.
The agent observes a reward $r_t$ for having selected $a_t$ at $s_t$.
In partially observable environments, the agent observes $o(s_{t},a_t,s_{t+1})\in\Omega$ drawn according to an observation function $\obsfunc:\sspace\times\aspace\to\Delta(\Omega)$.
The agent's goal is to find the policy that maximizes her expected discounted return.
Formally, by letting $R_t\defeq\sum_{i=0}^\infty \gamma^i r_{t+i}$, the optimal policy $\pi^\ast$ is such that $\pi^\ast\in\argmax\expe_\pi[R_0]$.

Value-based algorithms compute $\pi^\ast$ by estimating $v_\pi(s)\defeq \expe_{\pi}[R_t|S_t=s]$ (state-value function) or $Q_\pi(s,a)\defeq \expe_{\pi}[R_t|S_t=s,A_t=a]$ (action-value function) via temporal difference learning, and producing a series of $\epsilon$-greedy policies.

A popular alternative is employing policy gradient methods.
The main ides is to directly adjust the parameters $\theta$ of a differentiable and parametrized policy $\pi_\theta$ via gradient ascent on a score function $\scorefunc(\pi_\theta)$.
Let $\rho_\pi(s_t,a_t)$ denote the state-action marginals of the trajectory distributions induced by a policy $\pi(a_t|s_t)$.
Then, gradient of the score function may be rewritten as follows via the policy gradient theorem~\cite{sutton2000policy}:
$\nabla_\theta J(\theta)=\expe_{(s,a)\sim \rho_\pi}[\nabla_\theta \log\pi_\theta(a|s) Q^\pi(s,a)]$.
One could learn an approximation of the true action-value function by, \eg, temporal-difference learning, and alternates between \emph{policy evaluation} and \emph{policy improvement} using the value function to obtain a better policy~\cite{barto1983neuronlike, sutton2018reinforcement}. In this case, the policy is referred to as the \emph{actor}, and the value function as the \emph{critic} leading to a variety of actor-critic algorithms~\cite{haarnoja2018sac, lillicrap2015continuous, mnih2016asynchronous, peters2008reinforcement, schulman2017proximal}.

\subsection{Extensive-Form Games}

Extensive-form games (EFGs) are a model of sequential interaction involving a set of players $\pl$.
Exogenous stochasticity is represented via a {\em chance} player (also called {\em nature}) which selects actions with a fixed known probability distribution.
A history $h\in \hi$ is a sequence of actions from all players (\emph{chance} included) taken from the start of an episode.
In imperfect-information games, each player only observes her own information states.
For instance, in a poker game a player observes her own cards, but not those of the other players.
An \textbf{information state} $s_t$ of player $i$ comprises all the histories which are consistent with player $i$'s previous observations.
We focus on EFGs with perfect recall, \ie, et every $t$ each player has a perfect knowledge of the sequence $s_1^i,a_1^i,\ldots,s_t^i$.

There are two causes for information states: i) private information determined by the environment (\eg, hands in a poker game), and ii) limitations in the observability of other players' actions.
When all information states are caused by i), we say the game is \textbf{perfectly observable}.

A policy which maps information states to probability distributions is often referred to as a \textbf{behavioral strategy}.
Given a behavioral strategy profile $\pi=(\pi_i)_{i\in\pl}$, player $i$'s incentive to deviate from $\pi_i$ is quantified as $e_i(\pi)\defeq \max_{\pi'_i}\expe_{\pi'_i,\pi_{-i}} [R_{0,i}]-\expe_{\pi}[R_{0,i}]$, where $\pi_{-i}=(\pi_j)_{j\in\pl\setminus\{i\}}$.
The \textbf{exploitability} of $\pi$ is defined as $e(\pi)\defeq\frac{1}{|\pl|}\sum_{i\in\pl}e_i(\pi)$.
\footnote{In constant-sum games (\ie, games where $\sum_{i\in\pl}r_{i,t}=k$ for each $t$) the exploitability is defined as $e(\pi)=\frac{\sum_{i\in\pl}e_i(\pi)-k}{|\pl|}$.}
A strategy profile $\pi$ is a \textbf{Nash equilibrium} (NE)~\cite{Nash1951} if $e(\pi)=0$.

In principle, a player $i$ could adopt a deterministic policy $\sigma_i: s\to \A$, selecting a single action at each information state.
From a game-theoretic perspective, $\sigma_i$ is an action of the equivalent tabular (\ie, normal-form) representation of the EFG.
We refer to such actions as \textbf{plans}.
Let $\Sigma_i$ be the set of all plans of player $i$.
The size of $\Sigma_i$ grows exponentially in the number of information states.
Finally, a \textit{normal-form strategy} $x_i$ is a probability distribution over $\Sigma_i$.
We denote by $\mathcal{X}_i$ the set of the normal-form strategies of player $i$.

\section{Team's Coordination: A Game-Theoretic Perspective}


A \textbf{team} is a set of players sharing the same objectives~\cite{basilico2017team,VonStengelKoller97}.
To simplify the presentation, we focus on the case in which a team of two players (denoted by T1, T2) faces a single adversary (A) in a zero-sum game.
This happens, \eg, in the card-playing phase of Bridge, where a team of two players, called the “defenders”, plays against a third player, the “declarer”.

Team members are allowed to communicate before the beginning of the game, but are otherwise unable to communicate during the game, except via publicly-observed actions.
This joint planning phase gives team members an advantage: for instance, the team members could skew their strategies to use certain actions to signal about their state (for example, that they have particular cards). In other words, by having agreed on each member’s planned reaction under any possible circumstance of the game, information can be silently propagated in the clear, by simply observing public information.

A powerful, game-theoretic way to think about about ex ante coordination is through the notion of \textbf{coordination device}.
In the planning phase, before the game starts, team members can identify a set of joint plans.
Then, just before the play, the coordination device draws one of such plans according to an appropriate probability distribution, and team members will act as specified in the selected joint plan.

Let $\Sigma_T=\Sigma_{T1}\times\Sigma_{T2}$ and denote by $R_{t,T}$ the return of the team from time $t$.
The game is such that, for any $t$, $R_{t,A}=-R_{t,T}$.
An optimal ex ante coordinated strategy for the team is defined as follows (see also~\citet{celli2018computational,Farina&Celli2017}):

\begin{definition}[TMECor]\label{def:tmecor}
	A pair $\zeta=(\pi_A,x_T)$, with $x_T\in \Delta(\Sigma_{T})$, is a {\em team-maxmin equilibrium with coordination device} (TMECor) iff
	\[e_A(\zeta)\defeq \max_{\pi'_A}\E{\pi'_A,(\sigma_1,\sigma_2)\sim x_T}{R_{0,A}}-\E{\pi_A,(\sigma_1,\sigma_2)\sim x_T}{R_{0,A}}=0,\]
	and
	\[e_T(\zeta)\defeq \max_{(\sigma'_1,\sigma'_2)\in\Sigma_T}\E{\pi_A,(\sigma'_1,\sigma'_2)}{R_{0,T}}-\E{\pi_A,(\sigma_1,\sigma_2)\sim x_T}{R_{0,T}}=0.\]
\end{definition}

In an approximate version, $\epsilon$-TMECor, neither the team nor the opponent can gain more than $\epsilon$ by deviating from their strategy, assuming that the other does not deviate.

By sampling a recommendation from a joint probability distribution over $\Sigma_T$, the coordination device introduces a correlation between the strategies of the team members that is otherwise impossible to capture using behavioral strategies (\ie, decentralized policies).
Coordinated strategies (\ie, strategies in $\Delta(\Sigma_T)$) may lead to team's expected returns which are arbitrarily better than the expected returns achievable via behavioral strategies.
This is further illustrated by the following example.

\begin{example}\label{ex:coordination_game}
A team of two players (T1, T2) faces an adversary (A) who has the goal of minimizing team's utility.
Each player of the game has two available actions and has to select one without having observed the choice of the other players.
Team members receive a payoff of $K$ only if they both guess correctly the action taken by the adversary, and mimic that action (\eg, when A plays left, the team is rewarded $K$ only if T1 plays $\ell$ and T2 plays \textsf{L}).
Team's rewards are depicted in the leaves of the tree in Figure~\ref{fig:coordination_game}.

\begin{figure}[!h]
	\centering \begin{tikzpicture}[scale=.3]
		\colorlet{darkgreen}{green!50!black};
	    \tikzstyle{treenode} = [circle,black,ultra thick,minimum size=0.7cm,inner sep=0];
	    \tikzstyle{leafn} = [black,minimum size=0.5cm,inner sep=0mm];
	    \tikzstyle{chance} = [fill=white!80!red,draw=red];
	    \tikzstyle{p1} = [fill=white!80!violet,draw=violet];
	    \tikzstyle{p2} = [fill=white!80!darkgreen,draw=darkgreen];

	    \tikzstyle{con} = [draw=black, shorten <=1mm, shorten >=1mm, line width=0.5mm];
		\tikzstyle{highlight} = [line width=1mm,draw=red,shorten <=1mm, shorten >=1mm,];

		\node[treenode,chance] (A) at (0, 0) {\bf \fontsize{12}{0}\selectfont {A}};

		\fill[fill=black!7!white] (-7, -3.2) rectangle (7, -5.8);
		\node[treenode,p1] (B) at (-7, -4.5) {\bf \fontsize{12}{0}\selectfont {T1}};
		\node[treenode,p1] (C) at (7, -4.5) {\bf \fontsize{12}{0}\selectfont {T1}};

		\fill[fill=black!7!white] (-10, -7.7) rectangle (10.1, -10.3);
		\node[treenode,p2] (D) at (-10, -9) {\bf \fontsize{12}{0}\selectfont {T2}};
		\node[treenode,p2] (E) at (-4, -9) {\bf \fontsize{12}{0}\selectfont {T2}};
		\node[treenode,p2] (F) at (4, -9) {\bf \fontsize{12}{0}\selectfont {T2}};
		\node[treenode,p2] (G) at (10, -9) {\bf \fontsize{12}{0}\selectfont {T2}};

		\node[leafn] (H) at (-12, -14) {\fontsize{12}{0}\selectfont {$K$}};
		\node[leafn] (I) at (-8, -14) {\fontsize{12}{0}\selectfont {$0$}};
		\node[leafn] (L) at (-6, -14) {\fontsize{12}{0}\selectfont {$0$}};
		\node[leafn] (M) at (-2, -14) {\fontsize{12}{0}\selectfont {$0$}};
		\node[leafn] (N) at (2, -14) {\fontsize{12}{0}\selectfont {$0$}};
		\node[leafn] (O) at (6, -14) {\fontsize{12}{0}\selectfont {$0$}};
		\node[leafn] (P) at (8, -14) {\fontsize{12}{0}\selectfont {$0$}};
		\node[leafn] (Q) at (12, -14) {\fontsize{12}{0}\selectfont {$K$}};

		\path[con] (A) --node[above left]{} (B);
		\path[con] (A) --node[above right]{} (C);

		\path[highlight] (B) --node[yshift=1mm, left]{ $\ell$} (D);
		\path[con] (B) --node[yshift=1mm, right]{ $r$} (E);

		\path[con] (C) --node[yshift=1mm, left]{ $\ell$} (F);
		\path[highlight] (C) --node[yshift=1mm, right]{ $r$} (G);

		\path[highlight] (D) --node[yshift=1mm, left]{ \textsf{L}} (H);
		\path[con] (D) --node[yshift=1mm, right]{ \textsf{R}} (I);

		\path[con] (E) --node[yshift=1mm, left]{ \textsf{L}} (L);
		\path[con] (E) --node[yshift=1mm, right]{ \textsf{R}} (M);

		\path[con] (F) --node[yshift=1mm, left]{ \textsf{L}} (N);
		\path[con] (F) --node[yshift=1mm, right]{ \textsf{R}} (O);

		\path[con] (G) --node[yshift=1mm, left]{ \textsf{L}} (P);
		\path[highlight] (G) --node[yshift=1mm, right]{ \textsf{R}} (Q);

		\path[con,gray,dashed,dash pattern=on .3cm off .3cm on .3cm off .3cm] (B) -- (C);
		\path[con,gray,dashed,dash pattern=on .3cm off .3cm on .3cm off .3cm] (D) -- (E);
		\path[con,gray,dashed,dash pattern=on .3cm off .3cm on .3cm off .3cm] (E) -- (F);
		\path[con,gray,dashed,dash pattern=on .3cm off .3cm on .3cm off .3cm] (F) -- (G);
	\end{tikzpicture}
    \caption{Extensive-form game with a team}
    \label{fig:coordination_game}
\end{figure}
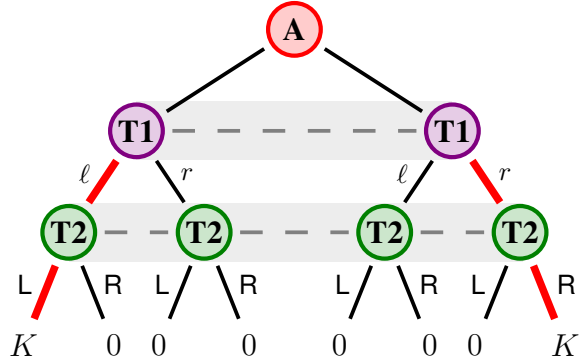

If team members didn't have the opportunity of communicating before the game, then the best they could do (\ie, the NE of the game) is randomizing between their available actions according to a  uniform probability.
This leads to an expected return for the team of $K/4$.

When coordinated strategies are employed, team members could decide to play with probability 1/2 the deterministic policies $\{\ell,\textsf{L}\}$ and $\{r,\textsf{R}\}$.
This is precisely the TMECor of the game, guaranteeing the team an expected return of $K/2$.
\end{example}

Classical multi-agent RL algorithms employ behavioral strategies and, therefore, are unable to attain the optimal coordinated expected return even in simple settings such as the one depicted in the previous example (see also Section~\ref{sec:exp}).
However, working in the space of joint plans $\Sigma_T$ is largely unpractical since its size grows exponentially in the size of the game instance.
In the following section we propose a technique to foster ex ante coordination without explicitly working over $\Sigma_T$.

\section{Soft Team Actor-Critic (STAC)}
\label{sec:method}

In this section, we describe Soft Team Actor-Critic (STAC), which is a scalable sample-based technique to approximate team's ex ante coordinated strategies.
Intuitively, STAC mimics the behavior of a coordination device via an exogenous signaling scheme.
The STAC's framework allows team members to associate shared meanings to initially uninformative signals.

First, we present a brief description of SAC~\cite{haarnoja2018sac} and a multi-agent adaptation of the original algorithm which may be employed to compute team's behavioral strategies.
Then, we describe a way to introduce a signaling scheme in this framework, which allows team members to play over ex ante coordinated strategies.

\subsection{Soft Actor Critic}
\label{sec:sac}
\textbf{Soft Actor Critic (SAC)}~\cite{haarnoja2018sac}, is an off-policy deep RL algorithm based on the maximum entropy (maxEnt) principle.
The algorithm exploits an actor-critic architecture with separate policy and value function networks.

The key characteristic of SAC is that the actor's goal is learning the policy which maximizes the expected reward while also maximizing its entropy in each visited state.
This is achieved via the following maximum entropy score function:
\begin{equation}\label{eq:sql:maxent_objective}
    \scorefunc(\pi)\defeq\sum_{t}\E{(s_t, a_t) \sim \rho_\policy}{\reward_t + \alpha\ent(\policy(\voidarg|s_t))},
\end{equation}
%
where $\alpha > 0$ is a temperature parameter that weights the importance of the entropy term.
In practice, stochastic policies are favored since the agent gains an extra reward proportional to the policy entropy.
This results in better exploration, preventing the policy from prematurely converging to a {\em bad} local optimum.
Moreover, SAC is designed to improve sample efficiency by reusing previously collected data stored in a replay buffer.

%

\subsection{Multi-Agent Soft Actor-Critic}
\label{sec:marl_sac}
SAC was originally designed for the single-agent setting. To ease training in multi-agent environments, we adopt the framework of \emph{centralized training with decentralized execution}, which allows team members to exploit extra information during training, as long as it is not exploited at test time.
This learning paradigm can be naturally applied to actor-critic policy gradient methods since they decouple the policy (\ie, the actor) of each agent from its $Q$-value function (\ie, the critic).
The actor has only access to the player's private state, which makes the policies decentralized.
This framework is a standard paradigm for multi-agent planning~\cite{kraemer2016multi, oliehoek2008optimal} and has been successfully employed by the multi-agent deep RL community to encourage the emergence of cooperative behaviors~\cite{foerster2016learning, foerster2018counterfactual, lowe2017multi}.

\paragraph{Actors} In imperfect-information games, team members are not homogeneous.
Intuitively, this is because: i) they have different private information, ii) they play at different decision points of the EFG.
Then, the sets of observations $\Omega_{T1}$, $\Omega_{T2}$ are disjoint.
This raises the need to have two separate policy networks, one for each team member, and optimize them separately~\cite{Chu&Ye2017}.

\paragraph{Critic}

In our setting, the critic has access to the \textbf{complete team state}, which contains the private information of each team member.
In the card game of Bridge, for instance, team members (actors) cannot observe each other cards.
However, at training time, the critic can exploit this additional information.
From a game-theoretic perspective, the critic works on an EFG with a finer information structure, such that if T1 and T2 were to be considered as a single player, the resulting player would have perfect recall.
Then, since team members' observations are shared, and since rewards are {\em homogeneous} by definition of team, we exploit {\em parameter sharing} and employ a single critic for the whole team.

\subsection{Signal Conditioning}

Ex ante coordinated strategies are defined over the exponentially large joint space of deterministic policies $\Sigma_T$.
Instead of working over $\Sigma_T$, STAC agents play according to decentralized policies conditioned on an exogenous signal, drawn just before the beginning of the game.
To do so, STAC introduces in the learning process an {\em approximate} coordination device, which is modeled as a fictitious player:

\begin{definition}[Signaler]\label{def:signaler}
	Given a set of signals $\Xi$ and a probability distribution $x_s\in\Delta(\Xi)$, a {\em signaler} is a non-strategic player which draws $\xi\sim x_s$ at the beginning of each episode, and subsequently communicates $\xi$ to team members.
\end{definition}

We assume that the number of signals is fixed, and that $x_s(\xi)=1/|\Xi|$ for all $\xi\in\Xi$.

At the beginning of the game, signals are completely {\em uninformative}, since they are drawn from a uniform probability distribution.
At training time, STAC allows team members for reaching a shared consensus about their joint behavior for each of the possible signals.
This is achieved as follows (see also Algorithm~\ref{alg:stac}):

\paragraph{Policy evaluation step}
STAC exploits a { \em value-conditioner network} whose parameters are produced via an {\em hypernetwork}~\cite{Ha2016}, conditioned on the observed signal $\xi$.
The role of the value-conditioner network is to perform action-value and state-value estimates by keeping $\xi$ into account.
This is a natural step since we expect team members to follow ad hoc policies for different observed signals.
In this phase, signal encodings are fixed.
%
%
Indeed, learning the hypernetworks' parameters has proven to worsen the performances of the architecture and, in some settings, prevented STAC from converging.

\paragraph{Policy Improvement Step}
The policy module is composed of a {\em policy conditioner network} whose parameters are produced by a fixed number of hypernetworks conditioned on the observed signal.
The policy module takes as input the local state of a team member, and outputs a probability distribution over the available actions.
The signals' hypernetworks are shared by all team members.
This demonstrated to be fundamental to associate a shared meaning to signals.
The policy conditioner net is updated by minimizing the expected Kullback-Leibler divergence as in the original SAC.

\begin{algorithm}[tb]
	\caption{Soft Team Actor-Critic}
	\label{alg:stac}
	\begin{algorithmic}[1]
		\Require $\params_1$, $\params_2$, $\pparams$, $\vparams$ \Comment{Initial parameters}
		\State $\vtargetparams\gets \vparams$ \Comment{Initialize target network weights}
		\State $\mathcal{D}\leftarrow\emptyset$ \Comment{Initialize an empty replay pool}
		\For{each iteration}

		\State $\xi \sim x_s$\Comment{The signaler draws $\xi$}\label{line:draw_signal}

		\For{each environment step}

		\State $\at \sim \policy_\pparams(\at|\st;\xi)$ \Comment{Sample action from the policy, conditioned on the signal}\label{line:new_action}
		\State $\stp \sim \transfunc(\st| \st, \at)$ \Comment{Sample transition from the environment}
		\State $\mathcal{D} \leftarrow \mathcal{D} \cup \left\{(\st, \at, \rt, \stp, \xi)\right\}$ \Comment{Store the transition in the replay pool}\label{line:sarsas}
		\EndFor

		\For{each gradient step}
		\State $\vparams\gets\vparams-\lambda_{V}\hat\nabla_{\vparams}J_V(\vparams)$
		\Comment{Update the $V$-function parameters}\label{line:update_vfunc}
		\State $\params_i \leftarrow \params_i - \lambda_Q \hat \nabla_{\params_i} J_\Q(\params_i)$ for $i\in\{1, 2\}$ \Comment{Update the $Q$-function parameters}\label{line:update_qfunc}
		\State $\pparams \leftarrow \pparams - \lambda_\policy \hat \nabla_\pparams J_\policy(\pparams)$\Comment{Update policy weights}\label{line:update_policy}
		\State $\vtargetparams\gets \tau \vparams + (1-\tau)\vtargetparams$ \Comment{Periodically update target network weights}\label{line:update_target}
		\EndFor
		\EndFor
	\end{algorithmic}
\end{algorithm}

\paragraph{Algorithm}
Algorithm~\ref{alg:stac} describes STAC's main steps.
STAC employs a parametrized state-value function $V_\vparams(\st;\xi)$, soft $Q$-function $Q_\qparams(\st,\at;\xi)$, and non-markovian policy $\pi_\pparams(\at|\st;\xi)$.
Each of this networks is conditioned on the observed signal $\xi$.

For each iteration, the signaler draws a signal $\xi\sim x_s$ (Line~\ref{line:draw_signal}).
Signals are sampled only before the beginning of each game.
Therefore, $\xi$ is kept constant for the entire trajectory (\ie, until a terminal state is reached).
Records stored in the replay buffer are augmented with the observed signal $\xi$ for the current trajectory.
We refer to these tuples (Line~\ref{line:sarsas}) as SARSAS records (\ie, a SARSA record with the addition of a signal).

After a fixed number of samples is collected and stored in the replay buffer, a batch is drawn from $\mathcal{D}$.
Since $x_s$ may be, in principle, an arbitrary distribution over $\Xi$, we decouple the sampling process from the choice of $x_s$ by ensuring the batch is composed by the same amount of samples for each of the available signals.
Then, parameters are updated as in SAC~\cite{haarnoja2018sac} (Lines~\ref{line:update_vfunc}-\ref{line:update_target}).
STAC employs a target value network $V_{\vtargetparams}(\st;\xi)$ to stabilize training~\cite{mnih2015human}.
In Line~\ref{line:update_qfunc}, STAC uses clipped double-$Q$ for learning the value function.
Specifically, taking the minimum $Q$-value between the two approximators has been shown to reduce positive bias in the policy improvement step~\cite{fujimoto2018addressing,van2016deep}.

\section{Experimental Evaluation}\label{sec:exp}

\subsection{Experimental Setting}
Our performance metric will be team's worst case payoff, \ie, team's expected payoff against a best-responding adversary.

\paragraph{Baselines} Neural Fictitious Self-Play (NFSP) is a sample-based variation of fictitious play introduced by~\citet{HeinrichS16}.
For each player $i$, a transition buffer is exploited by a DQN agent to approximate a best response against $\bar\pi_{-i}$ (\ie, the average strategies of the opponents).
A second transition buffer is used to store data via reservoir sampling, and is used to approximate $\pi_i$ by classification.
The first baseline ({\em NFSP-independent}) is an instantiation of an independent NFSP agent for each team member.
The second baseline ({\em NFSP-coordinated-payoffs}) is equivalent to NFSP-independent but, this time, team members observe \emph{coordinated payoffs} (\ie, they share their rewards and split the total in equal parts).
The last baseline is the multi-agent variation of SAC presented in Section~\ref{sec:marl_sac}, which allows us to assess the importance of the signaling framework on top of this architecture.

\paragraph{Game Instances}
We start by evaluating STAC on instances of the coordination game presented in Example~\ref{ex:coordination_game}.
We also consider the setting in which there's an imbalance in the team's payoffs, \ie, instead of receiving $K$ for playing both $\{\ell, L\}$ and $\{r, R\}$, team members receive $K/2$ and $K$, respectively.

Then, we consider a three-player Leduc poker game instance~\cite{southey2012bayes}, which is a common benchmark for imperfect-information games.
We employ an instance with rank $3$ and two suits of identically-ranked cards.
The game proceeds as follows: each player antes $1$ chip to play.
Then, there are two betting rounds, with bets limited to $2$ chips in the first, and $4$ chips in the second.
After the first betting round, a public card is revealed from the deck.
The reward to each player is the number of chips they had after the game, minus the number of chips they had before the game.
Team members' rewards are computed by summing rewards of T1 and T2, and then dividing the total in equal parts.

\paragraph{Architecture details}
Both STAC's critics and policies are parameterized with a multi-layer perceptron (MLP) with ReLU non-linearities, followed by a linear output layer that maps to the agents' actions. The number of hidden layers and neurons depend on the instance of the game. For the coordination game we used a single hidden layer and $64$ neurons, while for Leduc-Poker three hidden layers with $128$ neurons each. Conditioned on the external signal, the hypernetworks generate the weights and biases of the MLP's layers online, via independent linear transformations, and an additional MLP respectively. We used Adam optimizer~\cite{kingma2014adam} with learning rates of $10^{-2}$ for the policy network and $0.5\times 10^{-3}$ for the value netowrks. At each update iteration we sample a batch of $128$ trajectories from the buffer of size $10^5$.

\subsection{Main Experimental Results}

\begin{figure}
\begin{minipage}{.49\textwidth}
		\includegraphics[width=.9\textwidth]{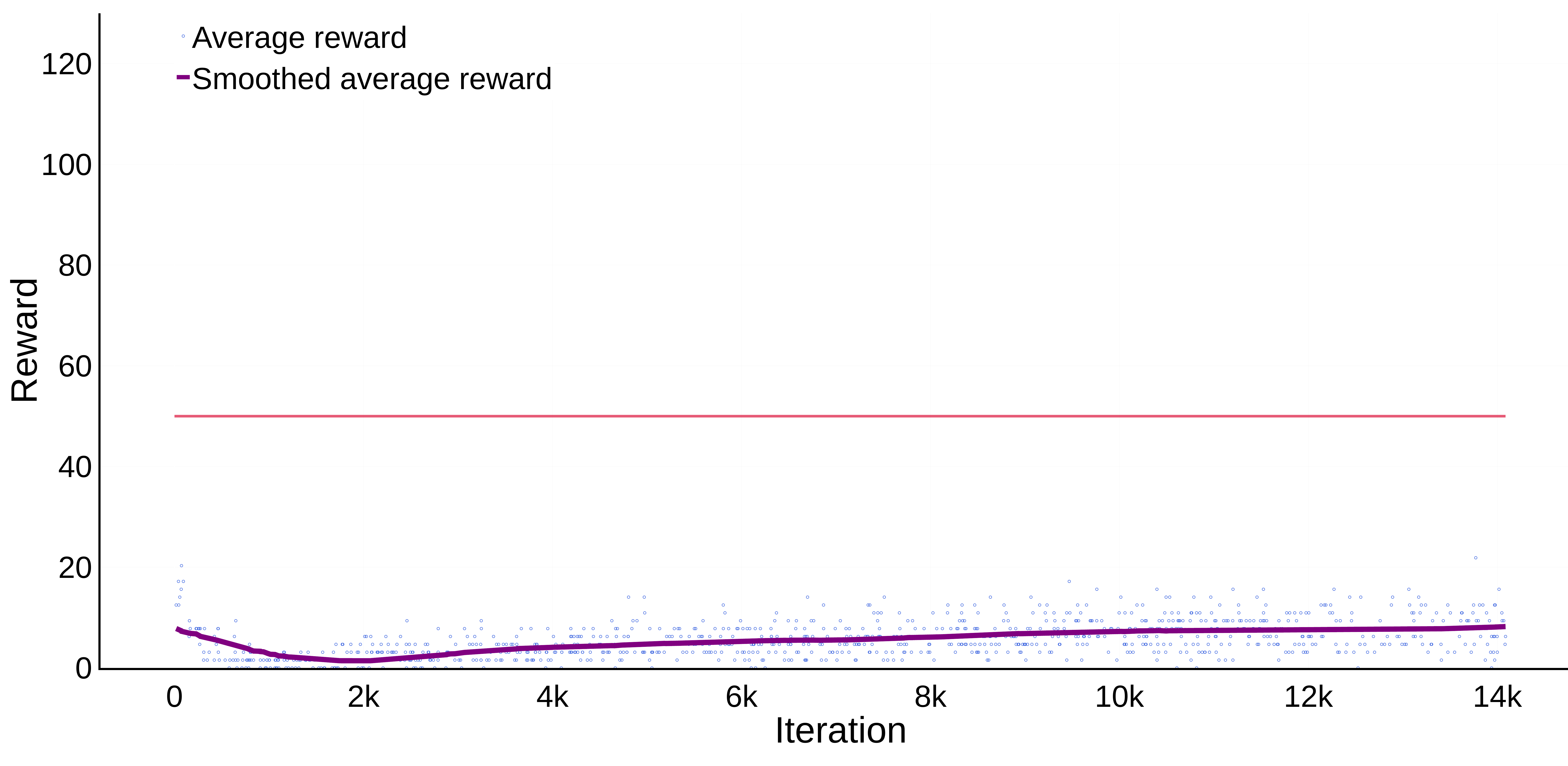}
\end{minipage}
\begin{minipage}{.49\textwidth}
		\includegraphics[width=.9\textwidth]{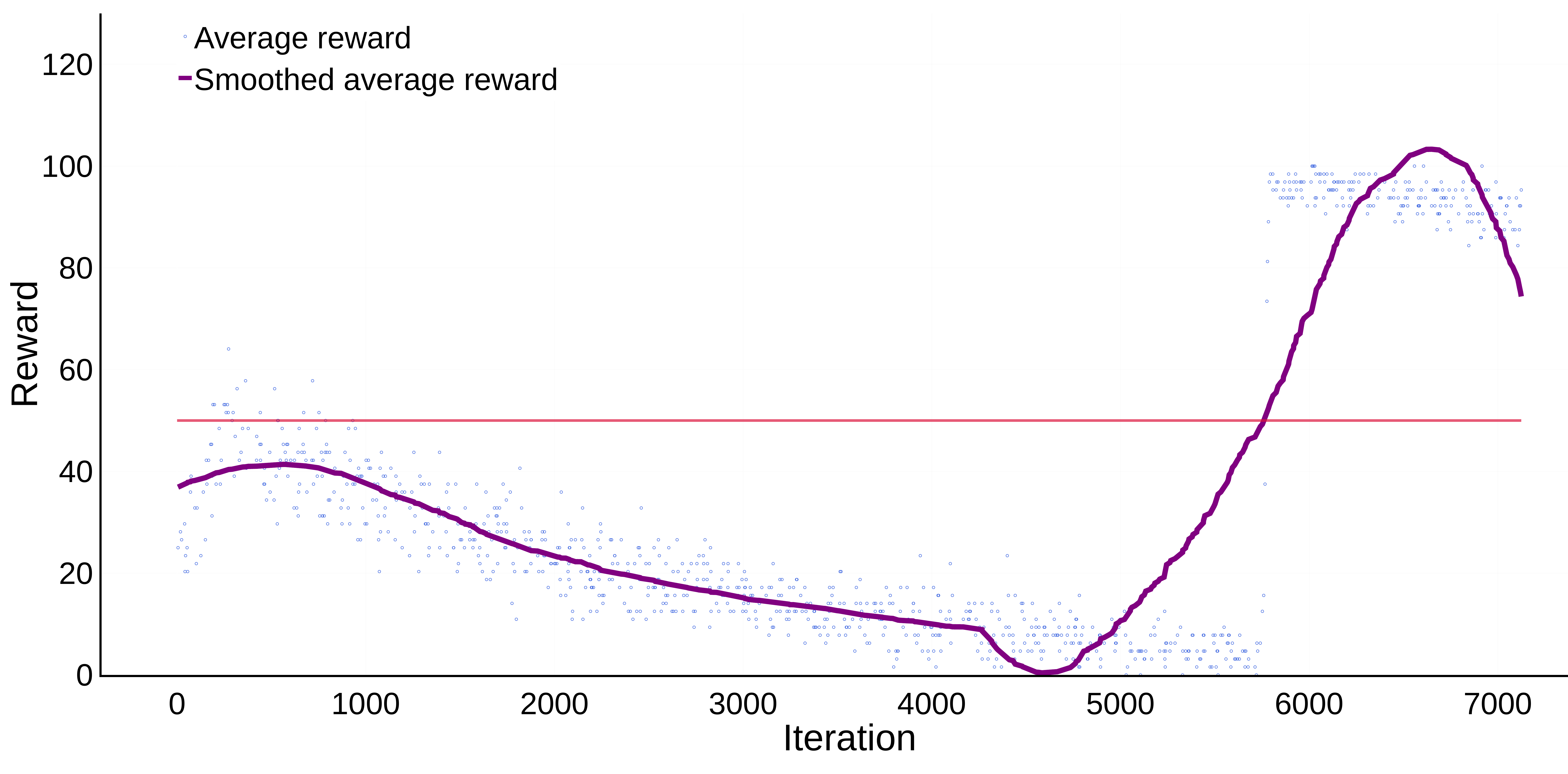}
\end{minipage}
\caption{Performances of the baseline algorithms on the coordination game of Example~\ref{ex:coordination_game} with $K=100$. The red line is the optimal worst-case expected payoff with coordination. {\em Left}: NFSP with coordinated payoofs. {\em Right}: Multi-agent SAC.}
\label{fig:coord_baselines}
\end{figure}

First, we focus on settings without perfect observability.
As expected, algorithms employing markovian policies fail to reach the optimal worst-case payoff with ex ante coordination (see Figure~\ref{fig:coord_baselines}).
NFSP-coordinated-payoffs is shown to guarantee a worst-case payoff close to $K/4$, which is the optimal value in absence of coordination.
Multi-agent SAC exhibits a cyclic behavior.
This is caused by team members alternating deterministic policies $\{\ell,L\}$, and $\{r,R\}$.
The team is easily exploited as soon as the adversary realizes that team members are always playing on the same side of the game.

\begin{figure}
	\begin{minipage}{.49\textwidth}
		\includegraphics[width=.9\textwidth]{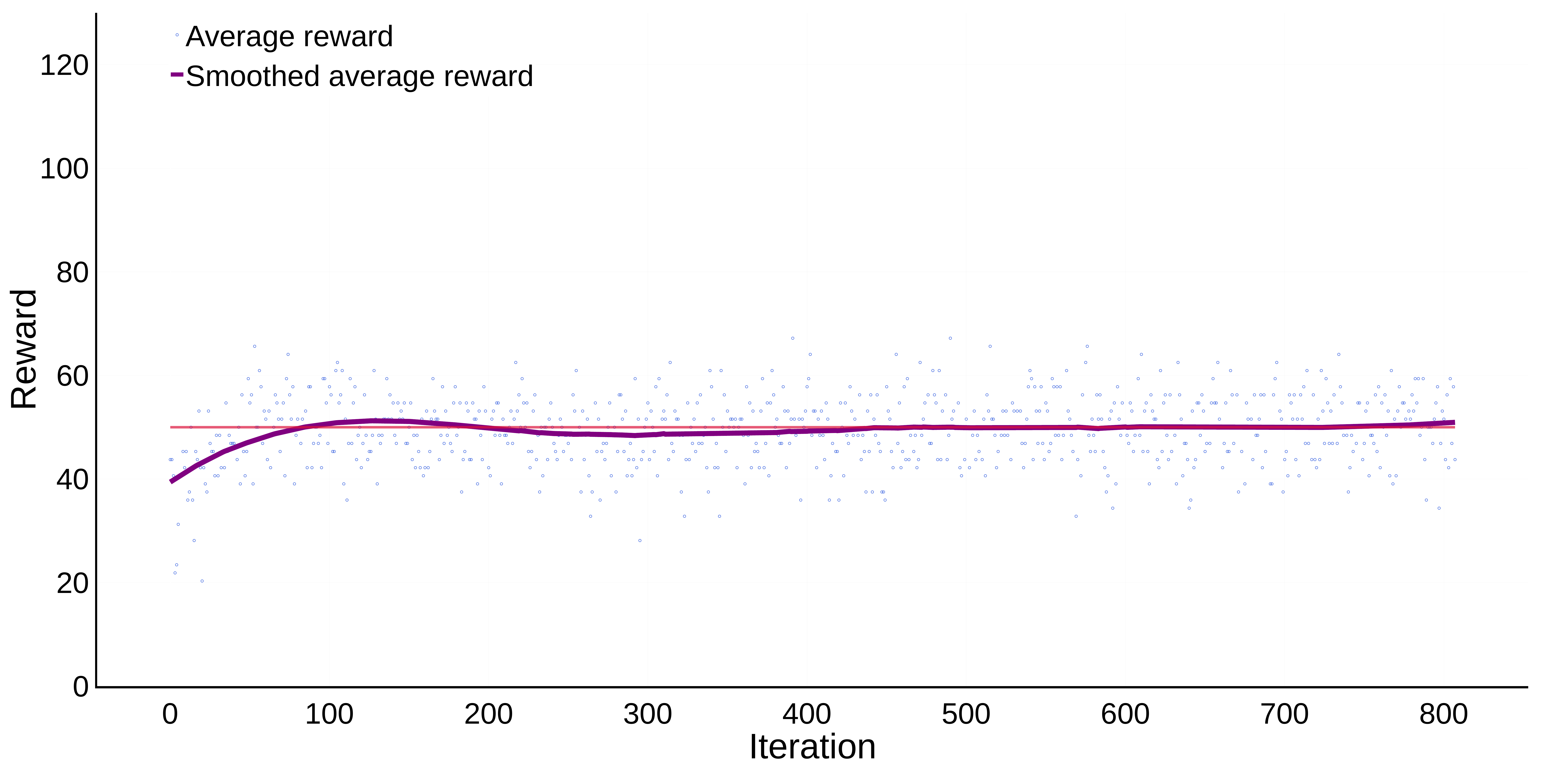}
	\end{minipage}
	\begin{minipage}{.49\textwidth}
		\includegraphics[width=.9\textwidth]{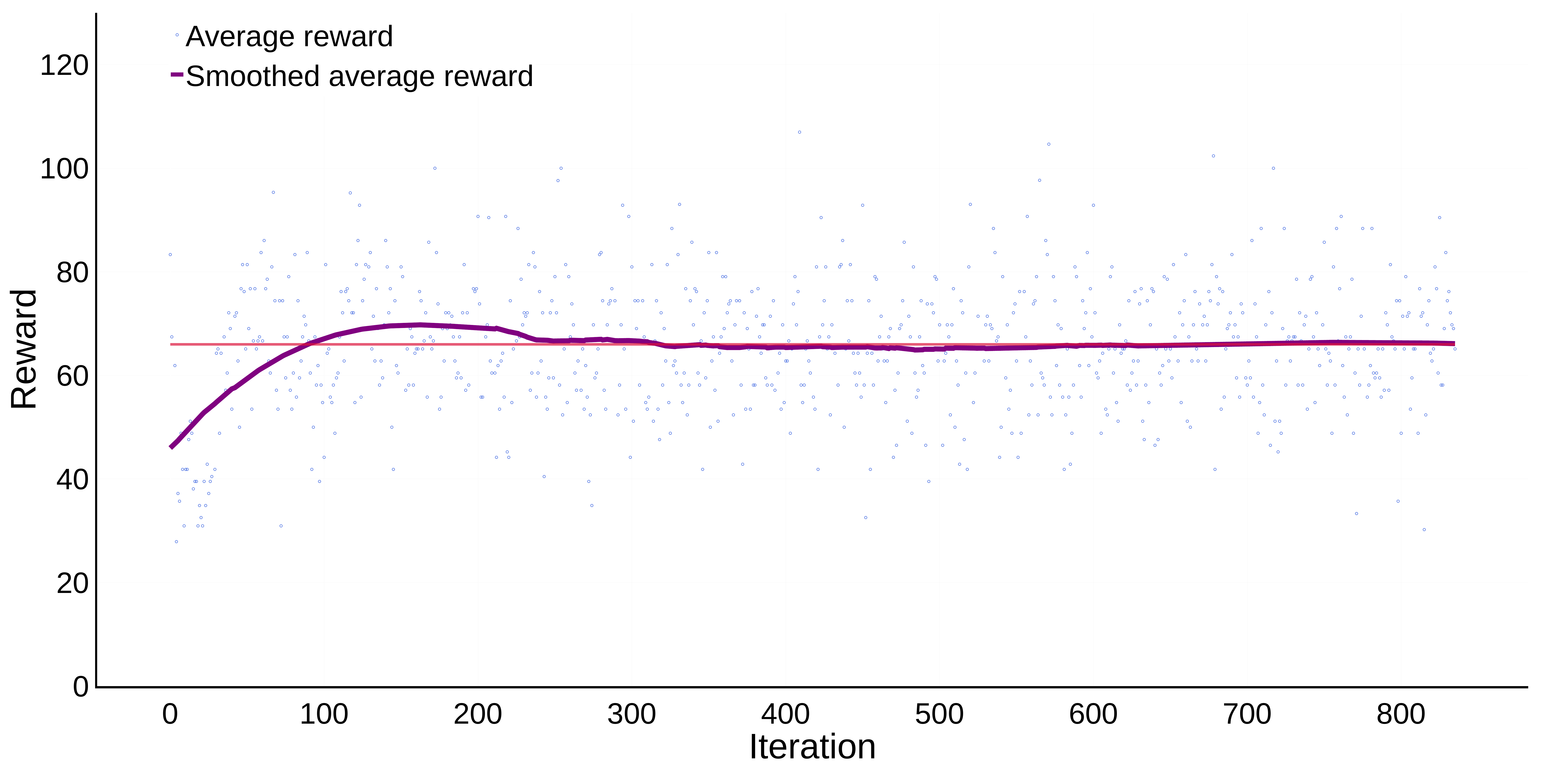}
	\end{minipage}
	\caption{Performances of STAC on different coordination games. {\em Left}: STAC with $|\Xi|=2$ on the balanced coordination game. {\em Right}: STAC with $|\Xi|=3$ on coordination game with $\{K,K/2\}$ strictly positive payoffs. Both games have $K=100$, and the optimal worst-case reward is marked with the red line.}
	\label{fig:coord_stac}
\end{figure}

Introducing signals allows team members for reaching near optimal worst-case expected payoffs (Figure~\ref{fig:coord_stac}).
The right plot of Figure~\ref{fig:coord_stac} shows STAC's performances on a coordination game structured as the one in Example~\ref{ex:coordination_game} but with team's strictly positive payoffs equal to $K$ and $K/2$.
In this case, a uniform $x_s$ over a binary signaling space cannot lead to the optimal coordinated behavior.
However, if an additional signal is added to $\Xi$ (\ie, $|\Xi|=3$), team members learn to play toward the terminal state with payoff $K/2$ after observing two out of three signals.
When the remaining signal is observed, team members play to reach the terminal state with payoff $K$.
This strategy is exactly the TMECor of the underlying EFG.
This suggests that even if $x_s$ is fixed to a uniform distribution, having a reasonable number of signals could allow good approximations of the optimal team's coordinated behaviors.

\begin{figure}
	\centering
	\includegraphics[width=.8\textwidth]{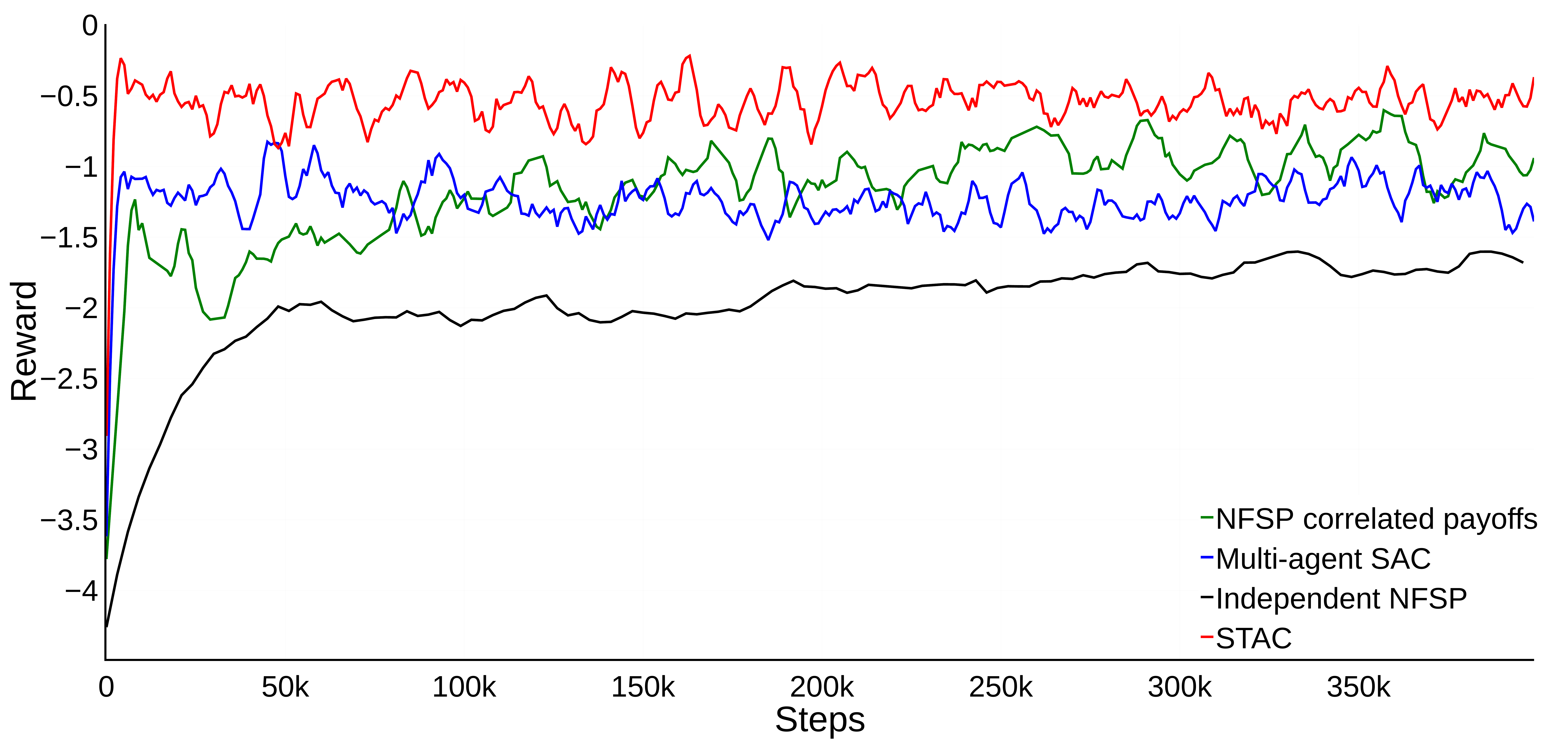}
	\caption{Comparison on the Leduc poker instance.}
	\label{fig:leduc}
\end{figure}

In perfectly observable settings such as Leduc poker, reaching a reasonable level of team coordination is less demanding.
We observed that most of the times it is enough to condition policies on the observed history of play to reach some form of implicit coordination among team members.
However, adding a signaling framework proved to be beneficial for team members even in this setting.
This can be observed in Figure~\ref{fig:leduc}.

\section{Related Works}
\label{sec:related_work}

Learning how to coordinate multiple independent agents~\cite{claus1998dynamics, boutilier1999sequential} via Reinforcement Learning requires tackling multiple concurrent challenges, \eg, non-stationarity, alter-exploration and shadowed-equilibria~\cite{matignon2012independent}. There is a rich literature of algorithms proposed for learning cooperative behaviours among independent learners. Most of them are based on heuristics encouraging agents' policies coordination~\cite{bloembergen2010lenient, bowling2001rational, lauer2000algorithm, lauer2004reinforcement, matignon2007hysteretic, matignon2008study, panait2005cooperative}.

Thanks to the recent successes of deep RL in single-agent environments~\cite{Mnih13, silver2016mastering, silver2018general}, MARL is recently experiencing a new wave of interest and some old ideas have been adapted to leverage the power of function approximators~\cite{omidshafiei2017deep, Palmer2018}. Several successful variants of the Actor-Critic framework based on the \emph{centralized training - decentralized execution} paradigm have been proposed~\cite{lowe2017multi, foerster2016learning, foerster2018counterfactual, sukhbaatar2016learning}. These works encourage the emergence of coordination and cooperation, by learning a centralized $Q$-function that exploits additional information available only during training. Other approaches factorize the shared value function into an additive decomposition of the individual values of the agents~\cite{Sunehag2017}, or combine them in a non-linear way~\cite{rashid2018qmix}, enforcing monotonic action-value functions. More recent works, showed the emergence of complex coordinated behaviours across team members in real-time games~\cite{jaderberg2019human, liu2018emergent}, even with a fully independent asynchronous learning, by employing population-based training~\cite{jaderberg2017population}.



In Game Theory, player's coordination is usually modeled via the notion of  \textit{Correlated equilibrium} (CE)~\cite{aumann1974subjectivity}, where agents make decisions following a recommendation function, \ie, a \emph{correlation device}.
Learning a CE of EFGs is a challenging problem as actions spaces grow exponentially in the size of the game tree.
A number of works in the MARL literature address this problem (see, \eg, ~\cite{cigler2011reaching, cigler2013decentralized, greenwald2003correlated, zhang2013coordinating}).
Differently from these works, we are interested in the computation of TMECor, as defined by~\cite{celli2018computational}.

In our work, we extend the SAC~\cite{haarnoja2018sac} algorithm to model the correlation device explicitly. 
By sampling a signal at the beginning of each episode, we show that the team members are capable to learn how to associate meaning to the uninformative signal.
Concurrently to our work,~\citet{chen2019signal} proposed a similar approach based on exogenous signals.
Coordination is incentivized by enforcing a mutual information regularization, while we make use of hypernetworks~\cite{Ha2016} to condition both the policy and the value networks to guide players towards learning a TMECor. 



%
%
%
%
%
%

\section{Discussion}



Equilibrium computation techniques for computing team's ex ante coordinated strategies usually assume a perfect knowledge of the EFG, which has to be represented explicitly~\cite{celli2018computational,Farina&Celli2017}.
We have introduced STAC, which is an end-to-end deep reinforcement learning approach to learning approximate TMECor in imperfect-information adversarial team games.
Unlike prior game-theoretic approaches to compute TMECor, STAC does not require any prior domain knowledge.

STAC fosters coordination via exogenous signals that are drawn before the beginning of the game.
Initially, signals are completely uninformative. 
However, team members are able to learn a shared meaning for each signal, associating specific behaviors to different signals.
Our preliminary experiments showed that STAC allows team members for reaching near optimal team's coordination even in settings that are not perfectly observable, where other deep RL methods failed to learn coordinated behaviors.
On the other hand, we observed that, in problems with perfect observability, a form of implicit coordination may be reached even without signaling, by conditioning policies on the observed history of play. 

\small
\bibliographystyle{plainnat}
\bibliography{bibliography.bib}

\end{document}